\documentclass{IEEEtran}
\usepackage{graphicx}
\usepackage{authblk}
\usepackage{hyperref}
\usepackage{subcaption}
\usepackage{booktabs}
\usepackage{cite}
\hypersetup{
    colorlinks=true,
    linkcolor=black,
    citecolor=black,
    filecolor=magenta,
    urlcolor=cyan,
    pdftitle={DeepSea MOT},
    pdfpagemode=FullScreen,
}
\title{DeepSea MOT: A benchmark dataset for multi-object tracking on deep-sea video}
\author[1]{Kevin Barnard}
\author[1,2]{Elaine Liu}
\author[1]{Kristine Walz}
\author[1]{Brian Schlining}
\author[1]{Nancy Jacobsen Stout}
\author[1]{Lonny Lundsten}
\affil[1]{Monterey Bay Aquarium Research Institute, 7700 Sandholdt Road, Moss Landing, CA 95039}
\affil[2]{University of Virginia, 1919 Ivy Road, Charlottesville, VA 22903}

\begin{document}

\maketitle

\begin{abstract}
    Benchmarking multi-object tracking and object detection model performance is an essential step in machine learning model development, as it allows researchers to evaluate model detection and tracker performance on human-generated `test' data, facilitating consistent comparisons between models and trackers and aiding performance optimization. In this study, a novel benchmark video dataset was developed and used to assess the performance of several Monterey Bay Aquarium Research Institute object detection models and a FathomNet single-class object detection model together with several trackers. The dataset consists of four video sequences representing midwater and benthic deep-sea habitats. Performance was evaluated using Higher Order Tracking Accuracy, a metric that balances detection, localization, and association accuracy. To the best of our knowledge, this is the first publicly available benchmark for multi-object tracking in deep-sea video footage. We provide the benchmark data, a clearly documented workflow for generating additional benchmark videos, as well as example Python notebooks for computing metrics.
\end{abstract}

\section{Introduction}
\label{sec:intro}

\begin{figure*}[ht!]
    \centering
    
    \begin{subfigure}{0.49\linewidth}
        \centering
        \includegraphics[width=\linewidth]{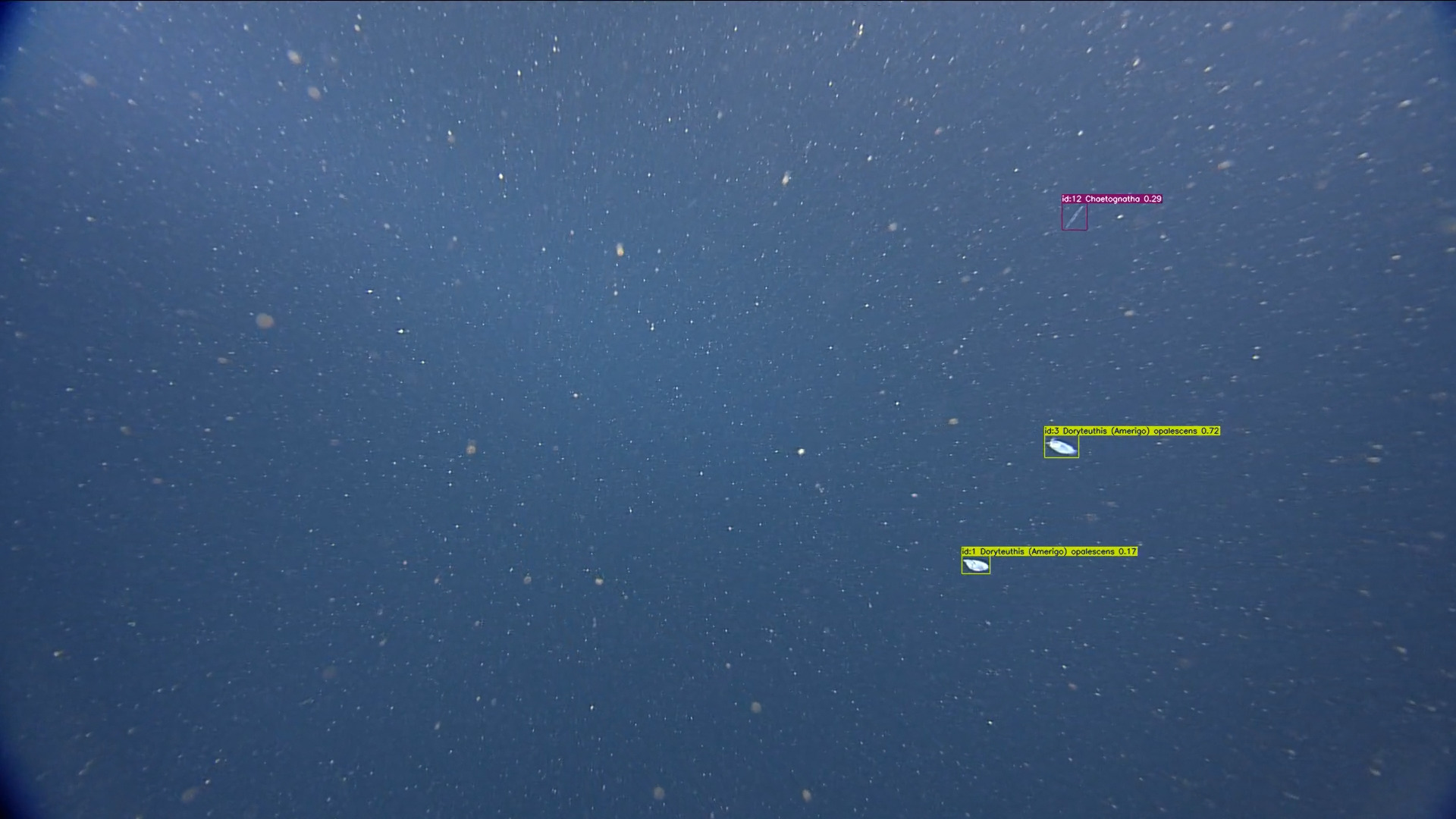}
        \caption{Midwater Simple (MWS).}
        \label{fig:seq-examples-a}
    \end{subfigure}
    \hfill
    \begin{subfigure}{0.49\linewidth}
        \centering
        \includegraphics[width=\linewidth]{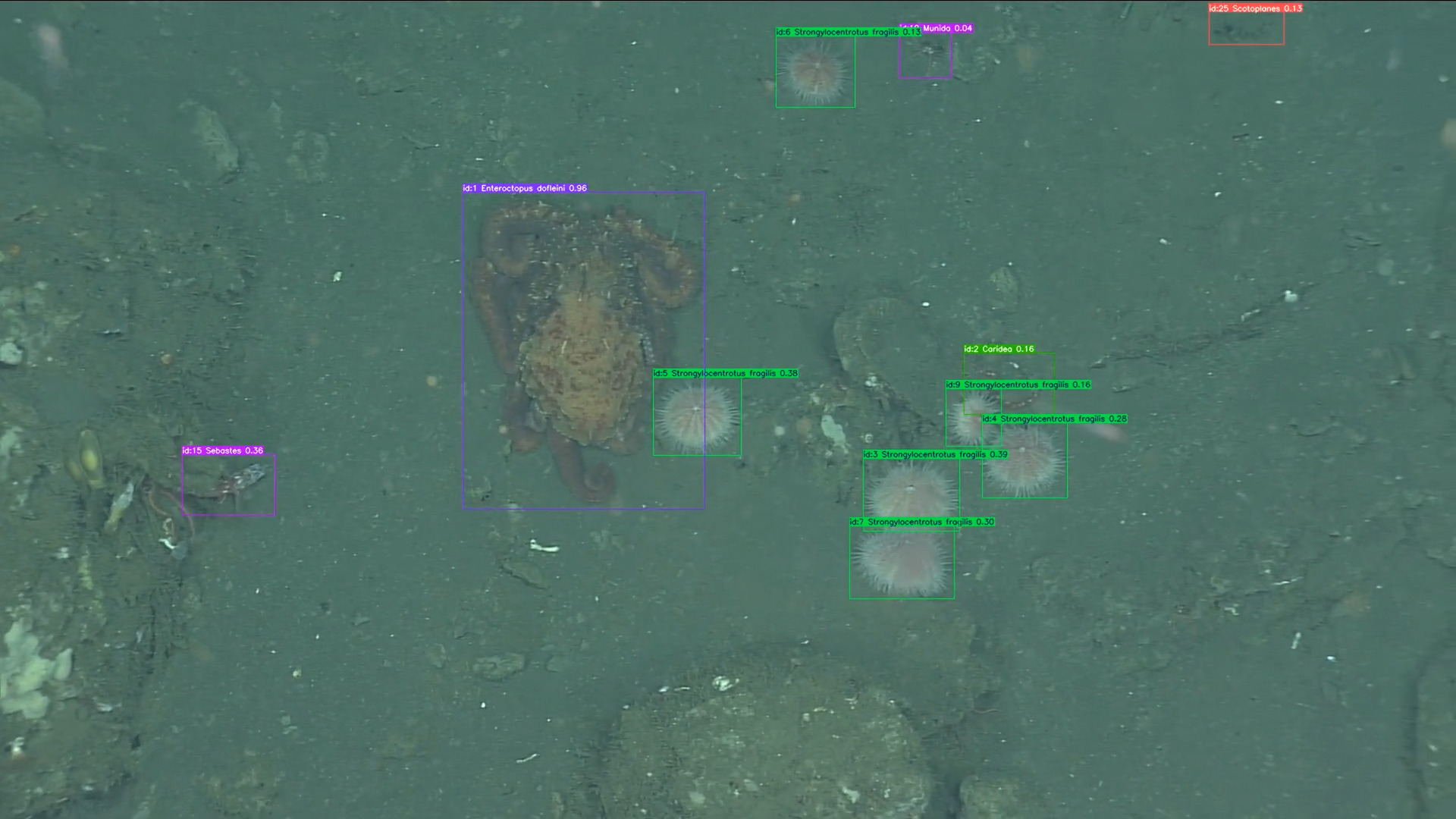}
        \caption{Benthic Simple (BS).}
        \label{fig:seq-examples-b}
    \end{subfigure}
    
    \vskip\baselineskip

    \begin{subfigure}{0.49\linewidth}
        \centering
        \includegraphics[width=\linewidth]{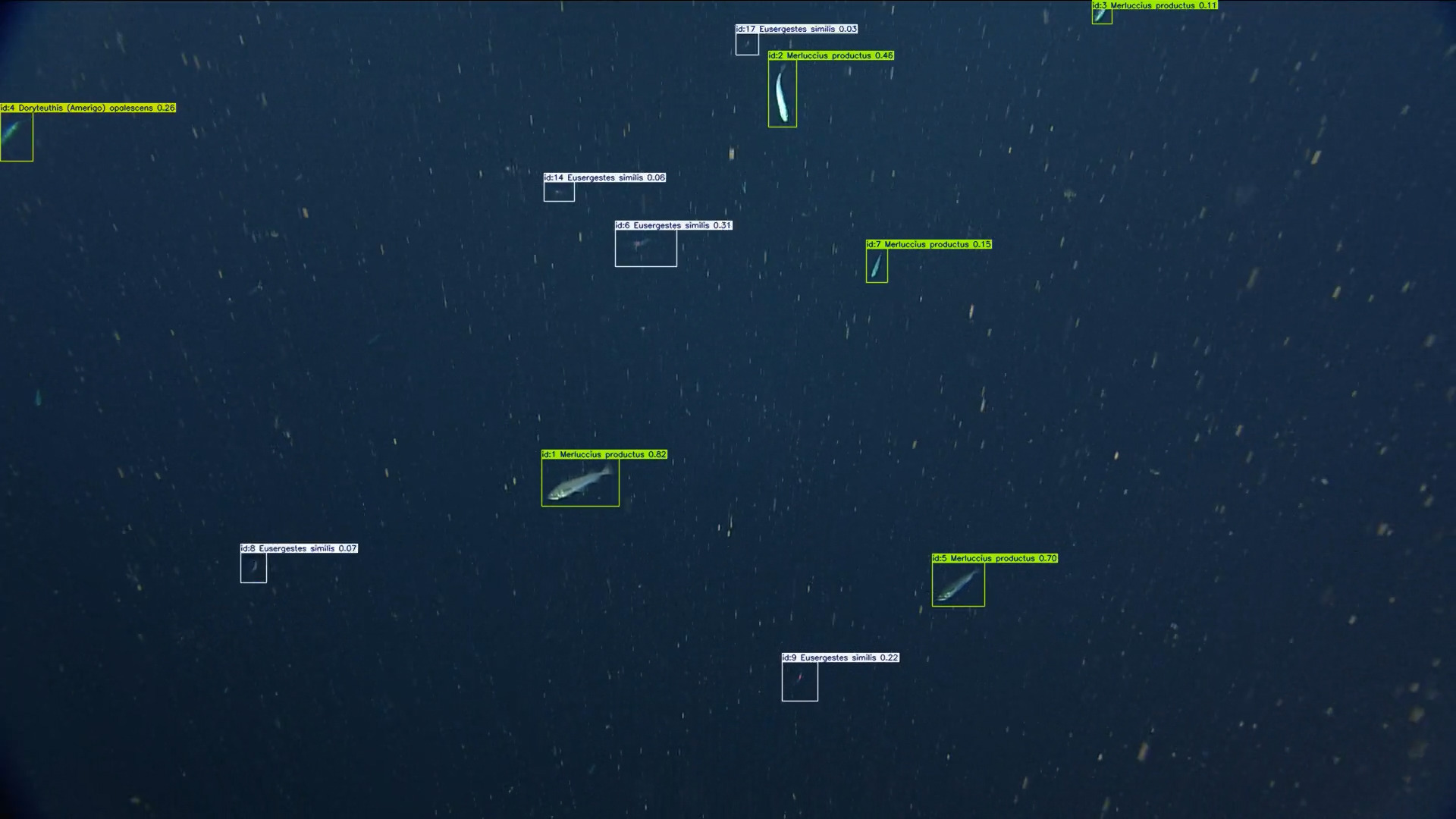}
        \caption{Midwater Difficult (MWD).}
        \label{fig:seq-examples-c}
    \end{subfigure}
    \hfill
    \begin{subfigure}{0.49\linewidth}
        \centering
        \includegraphics[width=\linewidth]{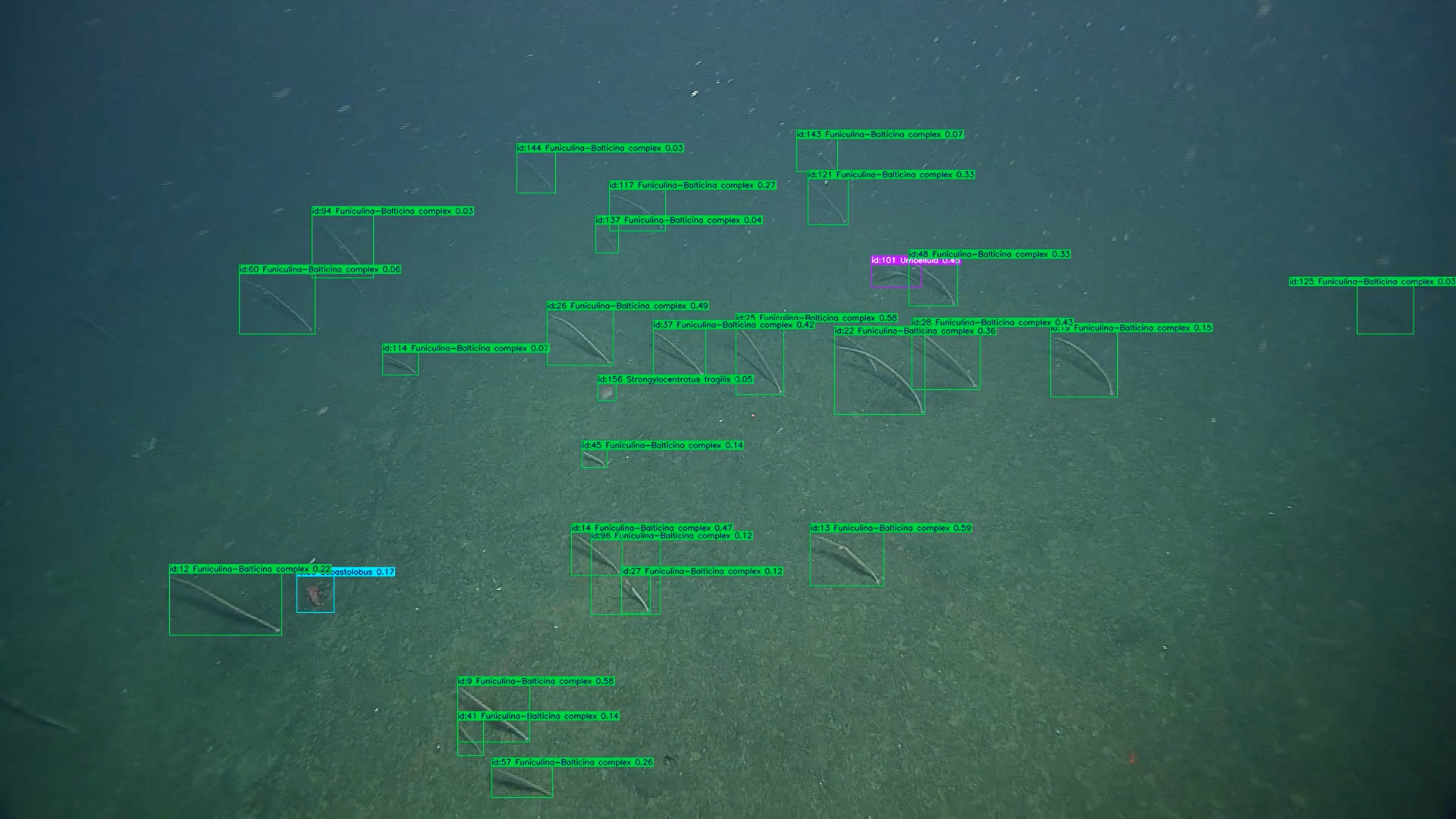}
        \caption{Benthic Difficult (BD).}
        \label{fig:seq-examples-d}
    \end{subfigure}
    
    \caption{Example frames from benthic and midwater benchmark videos.}
    \label{fig:seq-examples}
\end{figure*}

As the volume of video footage and still images being captured to study the world ocean continues to grow, so does the demand for efficient processing and analysis of these data. The Monterey Bay Aquarium Research Institute (MBARI) stores its digital video and image archive on a multi-petabyte, networked solid-state drive storage system managed by the MBARI Media Management system (VARS-M3, publication forthcoming). Observations gleaned from the visual media are archived within the Video Annotation and Reference System (VARS), a database with over 11 million annotations from 37 years of archived deep-sea imagery \cite{schlining2006mbari}. Annotating video and imagery is labor intensive and requires highly specialized domain expertise. For many organizations, the time and costs associated with doing this work is prohibitive. For example, one hour of video can take 2–3 hours or more to annotate manually. \cite{schlining2006mbari} describes the annotation process and VARS system in more detail. Machine learning (ML) offers a pathway to significantly reduce the time, effort, and costs associated with manual annotation. Optimized ML models can also support real-time object detection, offering the potential of at-sea ML assistance and add-on inferencing applications, such as detection, tracking, and ROV control as described in \cite{katija2021visual}.

Multi-object tracking (MOT) \cite{lealtaixe2015motchallenge} is made up of two main components: object detection and association. Object detection refers to what the object is (classification) and where the object is (localization). Association is a detected object's trajectory throughout the video, across the frames where it is observed. The object detection model handles object detection by generating classified bounding boxes in every single video frame. The tracker is responsible for association, and does so by linking detections together, from one frame to the next, correlating these detected objects. Thus, to successfully perform multi-object tracking, both the model and tracker need to be working accurately and cohesively. 

To evaluate MOT performance, a benchmark dataset made up of human-generated ground truth annotations is essential. Most existing MOT benchmarks have been developed for autonomous driving or human tracking and focus on pedestrians or vehicles \cite{zhang2023animaltrack}. Wild Animal Tracking Benchmark \cite{wu2015otb, fan2021totb} and AnimalTrack \cite{zhang2023animaltrack} provided some of the first dedicated multi-animal tracking benchmarks, but they primarily feature terrestrial animals. The AnimalTrack benchmark does contain aquatic animals like dolphins and penguins, but these are not representative of typical deep-sea communities. In addition, the deep-sea environment poses unique challenges: low visibility, near constant camera or animal motion, high biodiversity, high organism density, and significant occlusion. To address this knowledge gap, we created the first published benchmark dataset for multi-object tracking using deep-sea footage collected by MBARI's remotely operated vehicles (ROVs). This benchmark dataset provides MBARI's Video Lab with a standardized evaluation framework to assess object detector and tracker performance.

We developed benchmark video datasets for two primary marine habitats: midwater and benthic. Midwater refers to the open-ocean habitat from \textasciitilde100 m to many kilometers depth, but outside visual range of the seafloor \cite{robison1995light}. This habitat contains diverse animal communities with a large variety of unusual motions, along with ``marine snow,'' organic particulates derived from surface productivity. The benthic habitat refers to the seafloor and bottom-dwelling animals \cite{gage1991deep}. Both habitats host high biodiversity (number of species) and density (abundance of individuals) but are visually distinct. The midwater is a three-dimensional blue environment, devoid of substrate, with a near constant rain of marine snow. Midwater organisms are typically drifters or active swimmers and often exhibit transparency \cite{robison1995light}. Benthic habitats are characterized by substrate type (rock or sediment), slope, and communities of organisms that attach, burrow, crawl, or swim. Given these contrasts, our benchmark datasets were selected to account for the broader differences between midwater and benthic habitats, each of which poses distinct challenges for model performance (e.g., detecting transparent organisms against a uniform blue background versus distinguishing dense aggregations of benthic animals on rocky or sedimented substrates). 

\section{Materials and Methods}
\label{sec:materials}

\begin{table*}[t]
    \centering
    \begin{tabular}{c|c c c c c c c}
        \toprule
        \textbf{Name} & \textbf{FPS} & \textbf{Resolution} & \textbf{Length} & \textbf{Tracks} & \textbf{Boxes} & \textbf{Density} & \textbf{Unique Classes} \\
        \midrule
        MWS & 59.94 & 1920x1080 & 600 (00:10) & 8 & 1,953 & 3.3 & 6 \\
        MWD & 59.94 & 1920x1080 & 600 (00:10) & 57 & 12,835 & 21.4 & 11 \\
        BS & 59.94 & 1920x1080 & 600 (00:10) & 29 & 13,880 & 23.1 & 10 \\
        BD & 29.97 & 1920x1080 & 600 (00:20) & 94 & 28,708 & 47.8 & 4 \\
        \bottomrule
    \end{tabular}
    \medskip
    \caption{Summary statistics for each benchmark video sequence in the DeepSea MOT dataset.}
    \label{tab:seq-stats}
\end{table*}

MBARI's benchmark dataset was developed to evaluate combined model and tracker performance and it consists of four video sequences: two videos for the midwater (MW) and two videos for the benthic (B) environment (\autoref{tab:seq-stats}). Each environment includes one simple (MWS, BS) and one difficult (MWD, BD) video sequence, with difficulty defined largely by the number of animals present (\autoref{fig:seq-examples}). Most videos are 10 seconds long, have a frame resolution of 1920$\times$1080 and a frame rate of 59.94 fps, however, the difficult benthic (BD) video is 20 seconds long and 29.97 frames per second (1920$\times$1080 resolution). To generate the ground truth files, each video was first run through MBARI's 452k model and tuned ByteTrack tracker to produce preliminary ML-generated tracks. The results and corresponding video frames were then imported to RectLabel, an image annotation tool, to manually fit localizations to the object or create them when objects were not detected by the model. After all 600 frames were reviewed for each video sequence, the resulting ground truth files were then considered a gold-standard benchmark, available for comparison to the model output files for evaluation. 

\begin{table}[b]
    \centering
    \begin{tabular}{c|cccc}
        \toprule
        \textbf{Detection Model} & \textbf{MWS} & \textbf{MWD} & \textbf{BS} & \textbf{BD} \\
        \midrule
        MBARI 315k & 67.681 & 52.725 & 62.932 & \textbf{70.508} \\
        MBARI 452k & \textbf{76.776} & \textbf{61.262} & \textbf{67.446} & 70.202 \\
        FathomNet Megalodon & 68.483 & 43.497 & 67.600 & 68.739 \\
        \bottomrule
    \end{tabular}
    \medskip
    \caption{HOTA scores of the three object detection models evaluated on the four benchmark videos.}
    \label{tab:hota-scores}
\end{table}

Three MBARI fine-tuned object detection models were used in these experiments (\autoref{tab:hota-scores}). We utilized existing YOLOv8x \cite{yolov8_ultralytics} checkpoints from the Ultralytics Python library \cite{jocher2023ultralytics} for our transfer learning. Details for the training of these models can be provided upon request. For this series of experiments, we used inference hyperparameters that had been previously optimized for accuracy with MBARI's typical imagery from this unique environment. The parameters used for the detection model inference are given in Appendix \ref{app:detect-params}. The tracker used in this study is ByteTrack \cite{zhang2022bytetrack} as implemented by Ultralytics \cite{jocher2023ultralytics}. The tracker parameters were previously fine-tuned by an author (K. Walz) using extensive visual analysis and iterative testing of various combinations of parameter configurations. The default and tuned tracker parameters are provided in Appendix \ref{app:track-params}.

The evaluation metric chosen was the Higher Order Tracking Accuracy (HOTA) \cite{luiten2021hota}, which is an accuracy score that equally weights detection, localization, and association. Submetrics (DetA, AssA, LocA, precision, recall) calculated for the HOTA score provided additional insights. TrackEval \cite{luiten2020trackeval} was used to compute the HOTA metric and additional evaluation metrics for MOT on our benchmark videos using three existing MBARI models. Experiments were run in Python 3.10.13 with Ultralytics v8.3.155 and PyTorch 2.7.1 within a Jupyter Notebook, which is available in our open code repository (\url{https://github.com/mbari-org/benchmark_eval}).

\section{Results}
\label{sec:results}

\begin{figure}[ht]
    \centering
    \includegraphics[width=0.95\linewidth]{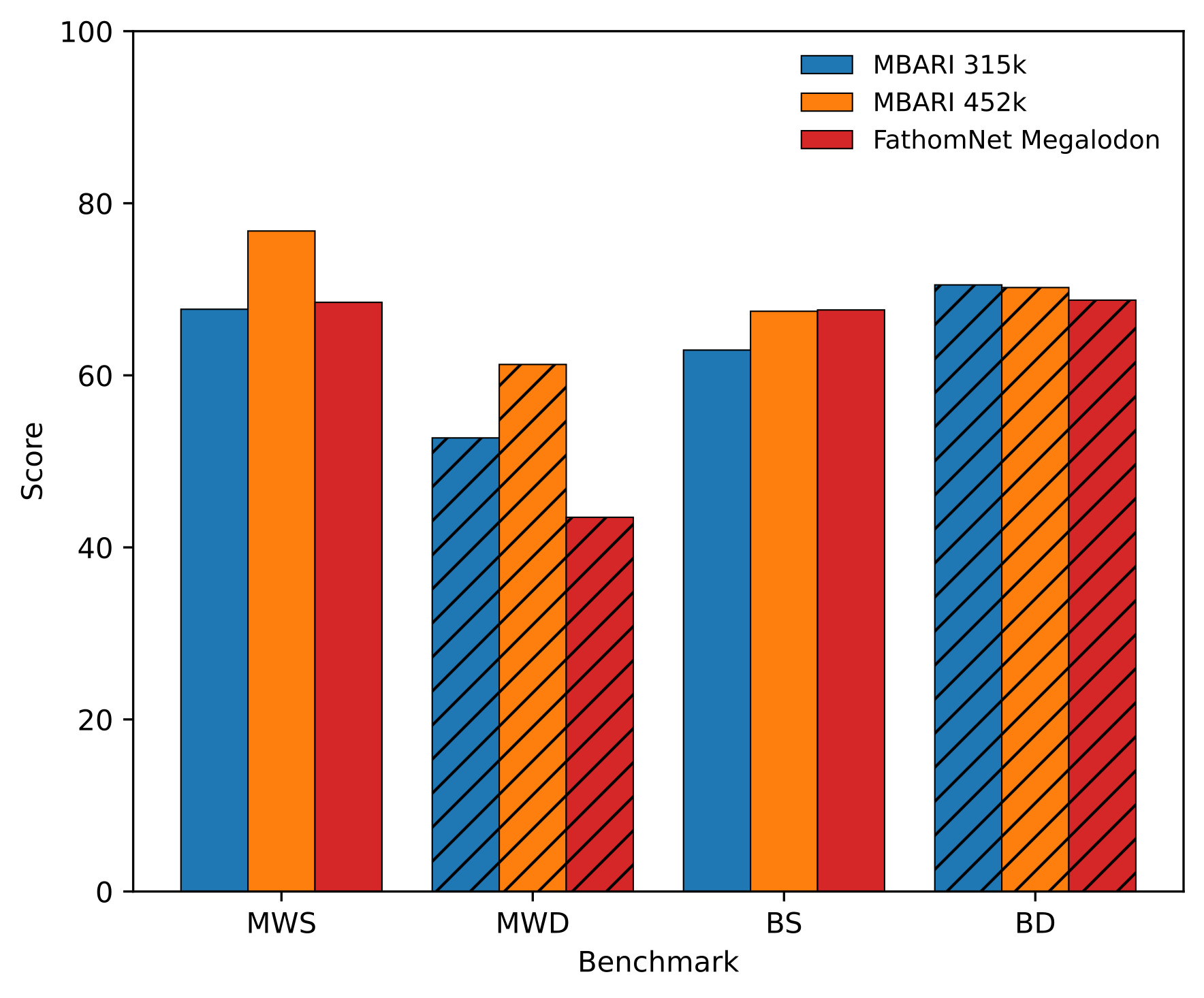}
    \caption{A comparison of HOTA scores for the three object detection models on four benchmark videos. Note that FathomNet Megalodon Detector is a single-class object detection model.}
    \label{fig:hota-comparison}
\end{figure}

\autoref{fig:hota-comparison} presents the HOTA scores for the four evaluated benchmark video sequences and three models used in this experiment. The highest HOTA score in this experiment was achieved using the MBARI 452k model with tuned ByteTrack on MWS, resulting in a score of $76.776\%$. The MBARI 452k model and tuned ByteTrack tracker achieved the best performance on all benchmarks except BD, where MBARI 315k garnered a score of $70.508\%$. MBARI 452k scored slightly lower on BD, at $70.202\%$. This result is in line with expectations\textemdash the major change between MBARI 452k and 315k was an increase in the amount of MW data provided for training the MBARI 452k model. Between MBARI 315k and MBARI 452k, benthic training data was largely unchanged, especially for the few taxa observed in BD. In general, the scores achieved were consistent with our initial assumption that HOTA scores would be higher on simpler video, as there were only a few animals to detect and objects were highly visible. The lowest performance was observed on MWD for all models. The lowest score, $43.497\%$, was achieved using the FathomNet Megalodon Detector (FNMD) \cite{fathomnet_megalodon}. This was also anticipated due to high occlusion, distant objects, and quick movement of the camera while the ROV descended through the midwater depths on the MWD benchmark. In particular, animals such as \emph{Poeobius meseres}, a small pelagic polychaete worm, were often missed because they appeared in the video as subtle shadows barely visible in the distant background. We were encouraged by the scores of MWS and MWD when comparing MBARI 315k to MBARI 452k as a great deal of effort was placed on increasing the number and quality of midwater localizations in the training data for MBARI 452k. This effort yielded a nearly $10\%$ increase in HOTA over the previous MBARI 315k model on MW benchmarks. We also achieved almost a $5\%$ increase in HOTA on BS when comparing MBARI 315k to MBARI 452k.

\begin{figure}[ht]
    \centering
    \includegraphics[width=0.95\linewidth]{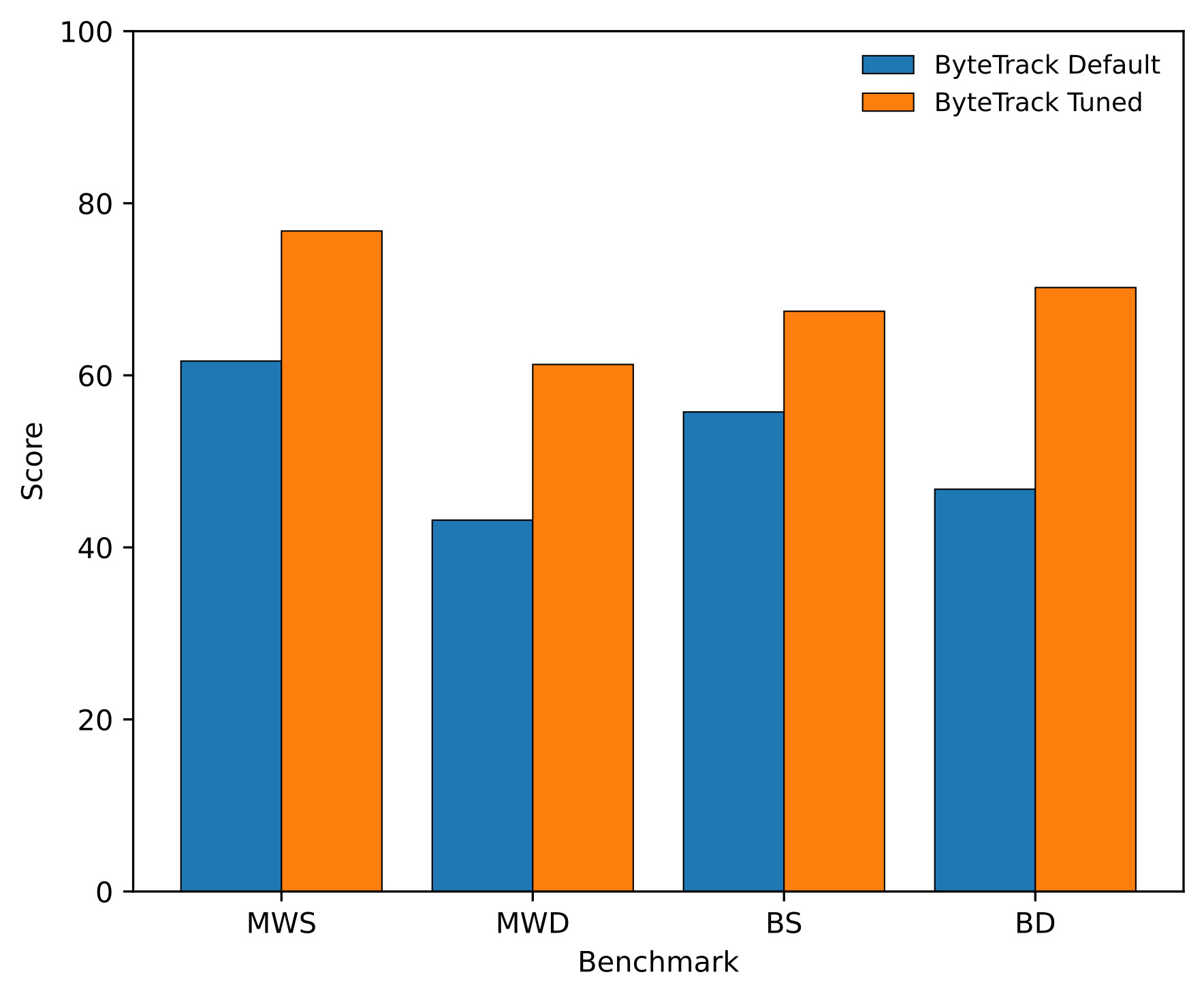}
    \caption{A comparison of HOTA scores for four benchmark videos before and after ByteTrack tracker parameter tuning.}
    \label{fig:hota-tuning}
\end{figure}

The untuned tracker with the default settings from Ultralytics showed much lower performance on average—approximately $15\%$ less compared to the same tracker with tuned hyperparameters (\autoref{fig:hota-tuning}). As noted previously, ByteTrack was tuned manually by careful visual review and iterative hyperparameter adjustments. In this case, metrics were used to report the results but were not used to tune the hyperparameters. The significant improvement in HOTA scores highlights the value of using domain knowledge and careful review of results when optimizing model and tracker performance.

Following model and tracker evaluation, an attempt was made to further optimize tracking using information gathered from the HOTA metric. The best performing tracker, ByteTrack (implemented through Ultralytics), was compared against BoT-SORT \cite{aharon2022botsort}, a tracker also supported by Ultralytics. Hyperparameter configurations were tested on the difficult midwater video sequence and parameters that were adjusted include the thresholds for first and second-stage associations, new track initialization, matching tracks, etc. Despite using supplemental information gathered from HOTA submetrics—such as recall and precision for detection and association—no improvements in accuracy were achieved, suggesting that the tracker was already highly optimized. This is supported by the significant differences observed in accuracy between the default and tuned tracker (\autoref{fig:hota-tuning}). 

In addition to the provided Jupyter notebooks on GitHub, comprehensive user documentation was created to support future development (\url{https://docs.mbari.org/benchmark_eval}). We hope the documentation will provide users with the ability to easily create additional benchmark videos specific to their own needs. Using benchmark videos, users can generate metrics when implementing new object detection models and trackers in an effort to improve performance on their data, using a model of their choosing. Links to the code, benchmark data, and documentation, all publicly available, are provided below. Documentation includes instructions for installation, usage, and helpful tips such as an explanation of directory file structure and file examples. 

\section{Discussion}
\label{sec:discussion}

These experiments and resulting HOTA metrics have led us to several conclusions pertaining to expectations of model performance with regard to video quality and underlying data used in model training. First, our HOTA scores suggest that the quality or content of the video (occlusion, motion, animal density) play a greater role in detection and tracking performance than habitat (benthic vs. midwater). For example, in the simple benthic video, the camera initially remains stationary, then pans to the side before returning to the original location. This movement caused the tracker to assign new IDs to the same animals when they temporarily left the frame, which largely reduced the HOTA score. On the other hand, the difficult benthic video has more animals present, but these are predominantly sea pens, which have minimal movement and are easier to track. Combined, these factors led to difficult benthic video achieving higher accuracy, underscoring the significance of video quality or content on model performance (\autoref{tab:hota-scores}). In addition, the quality and completeness of the underlying model training data have a huge impact on the resulting performance of the detection and tracking system. For example, significant effort was placed on incrementally improving our detection model performance, especially on midwater taxa. The results of this effort are noted in the significant increase in HOTA scores for the midwater benchmarks when comparing the MBARI 315k model to the MBARI 452k model (\autoref{tab:hota-scores}).

In future work, we intend to implement additional, state-of-the-art models and trackers in an attempt to improve overall performance. In addition to the trackers implemented in Ultralytics, we also ran experiments using BoxMOT \cite{brostrom2023boxmot}, an open source pluggable Python MOT tracker library, which included a range of high performance trackers such as BoostTrack \cite{stanojevic2024boosttrack}, StrongSORT \cite{du2023strongsort}, OC-SORT \cite{cao2023observation}, Deep OC-SORT \cite{maggiolino2023deep}, ByteTrack, and BoT-SORT. However, after evaluation, HOTA scores were significantly lower, even for ByteTrack and BoT-SORT. Reasons for this discrepancy are unclear, but could be due to differences in internal tracker implementations. 

We also intend on expanding the benchmark dataset to incorporate video sequences with greater diversity and complexity, which would enable a well-rounded evaluation of model performance on a more varied set of benchmark data. As these are generated, we will continue to update our benchmark data archive on Hugging Face (\url{https://huggingface.co/datasets/MBARI-org/DeepSea-MOT}).

\section{Conclusion}
\label{sec:conclusion}

By developing the first published benchmark dataset for multi-object tracking on deep-sea video footage, we aim to provide a foundation for systematic evaluation of deep-sea MOT. The dataset includes simple and difficult sequences from benthic and midwater environments. Evaluation with HOTA showed that tracker hyperparameter tuning has a greater impact than environmental setting, but that video quality can greatly impact overall object detection and tracker performance. Training dataset curation and hyperparameter tuning guided by careful visual review significantly boosted performance. Importantly, domain expertise remains critical for interpreting results and guiding improvements. Future efforts include expanding our benchmark datasets to longer and more diverse video sequences. We hope this work will be of great value when assessing detector model and tracker improvements moving forward, and provide useful metrics to guide selection and implementation of new tools in this rapidly evolving space.

\section*{Acknowledgments}

We thank the MBARI Video Lab and VARS teams. We acknowledge the captain and crew of the R/Vs \emph{Western Flyer} and \emph{Rachel Carson} and the pilots of the ROVs \emph{Doc Ricketts} and \emph{Ventana} for at-sea support. We thank Kyra Schlining for keen-eyed copyediting. Funding was provided by the Dean and Helen Witter Family Fund, the Rentschler Family Fund, and the Maxwell/Hanrahan Foundation. We are grateful to the David and Lucile Packard Foundation for continued support of MBARI.

\medskip

\bibliography{sources}

\appendices

\section{Detection Parameters}
\label{app:detect-params}

The following table gives the non-default detection parameters used for each of the three models evaluated in \autoref{sec:materials} per the implementation provided by \cite{jocher2023ultralytics}.

\medskip

\begin{center}
    \begin{tabular}{c|ccc}
        \textbf{Parameter} & \textbf{315k} & \textbf{452k} & \textbf{FNMD} \\ \hline
        \texttt{agnostic\_nms} & \texttt{True} & \texttt{True} & \texttt{True} \\
        \texttt{imgsz} & \texttt{1280} & \texttt{1280} & \texttt{1280} \\
        \texttt{conf} & \texttt{0.001} & \texttt{0.001} & \texttt{0.15} \\
        \texttt{iou} & \texttt{0.2} & \texttt{0.2} & \texttt{0.6} \\
        \texttt{augment} & \texttt{True} & \texttt{True} & \texttt{True} \\
    \end{tabular}
\end{center}

\section{Tracking Parameters}
\label{app:track-params}

The following table gives the ByteTrack parameters used for the evaluation experiments in \autoref{sec:materials}. Default values are provided per the implementation provided by \cite{jocher2023ultralytics}.

\medskip

\begin{center}
    \begin{tabular}{c|cc}
        \textbf{Parameter} & \textbf{Default} & \textbf{Tuned} \\ \hline
        \texttt{track\_high\_thresh} & \texttt{0.25} & \texttt{0.01} \\
        \texttt{track\_low\_thresh} & \texttt{0.10} & \texttt{0.005} \\
        \texttt{new\_track\_thresh} & \texttt{0.25} & \texttt{0.001} \\
        \texttt{track\_buffer} & \texttt{30} & \texttt{300} \\
        \texttt{match\_thresh} & \texttt{0.80} & \texttt{0.80} \\
        \texttt{fuse\_score} & \texttt{True} & \texttt{False} \\
    \end{tabular}
\end{center}

\end{document}